%% file: MLG-ECML-2024-paper.tex
\documentclass[runningheads]{llncs}

\input{src/preamble.tex}

\begin{document}

\input{src/title.tex}

\input{src/abstract.tex}

\input{src/introduction.tex}
\input{src/problem-statement.tex}

\input{src/related-work.tex}
\input{src/our-method.tex}
\input{src/experimental-evaluation.tex}
\input{src/conclusion.tex}

\input{src/credits.tex}

\bibliography{citations}

\end{document}

%% file: src/preamble.tex
\usepackage{adjustbox}
\usepackage{algorithm}
\usepackage{algpseudocode}
\usepackage{amsfonts}
\usepackage{amsmath}
\usepackage{bm}
\usepackage{booktabs}
\usepackage[utf8]{inputenc}
\usepackage{csquotes}
\usepackage{geometry}
\usepackage{hyperref}
\usepackage[misc]{ifsym}
\usepackage{todonotes}
\usepackage{listings}
\usepackage{color}
\definecolor{gray}{rgb}{0.5,0.5,0.5}
\definecolor{mauve}{rgb}{0.58,0,0.82}
\hypersetup{
	pdfencoding=auto,
	unicode=true,
	bookmarksopen=true,
	bookmarksopenlevel=3
}

\newcommand{\corr}{(\Letter)}
\newcommand{\name}[1]{\textit{#1}}
\newcommand{\mathfield}{\ensuremath{\mathbb}}
\newcommand{\mathmat}{\ensuremath{\mathbf}}

\newcommand{\mathvec}{\ensuremath{\bm}}

\newcommand{\method}{CSP}
\newcommand{\methodlong}{Convolutional Signal Propagation}
\newcommand{\U}{V} 
\newcommand{\uu}{v} 
\newcommand{\V}{E} 
\newcommand{\vv}{e} 
\newcommand{\HG}{\mathcal{H}} 
\newcommand{\HH}{\mathmat{H}} 
\newcommand{\BG}{\mathcal{G}_\mathrm{bip}} 
\newcommand{\E}{E_\mathrm{bip}} 
\newcommand{\D}{\mathmat{D}_v} 
\newcommand{\B}{\mathmat{D}_e} 
\newcommand{\X}{\mathmat{X}} 
\newcommand{\Y}{\mathmat{Y}} 
\newcommand{\y}{y} 
\newcommand{\x}{\mathvec{x}} 
\newcommand{\Tr}{V_\mathrm{train}} 
\newcommand{\vdeg}{\mathrm{d}} 
\newcommand{\edeg}{\delta} 

%% file: src/title.tex
\title{Convolutional Signal Propagation: A Simple Scalable Algorithm for Hypergraphs}
\titlerunning{Convolutional Signal Propagation}

\author{
	Pavel Procházka\inst{1} \and
	Marek Dědič\inst{1 2} \corr \orcidID{0000-0003-1021-8428} \and
	Lukáš Bajer\inst{1} \orcidID{0000-0002-9402-6417}
}
\authorrunning{P. Procházka et al.}

\institute{
	Cisco Systems, Inc., Karlovo náměstí 10, Prague, 120 00, Czech Republic \email{\{paprocha,madedic,lubajer\}@cisco.com} \and
	Faculty of Nuclear Sciences and Physical Engineering, Czech Technical University in Prague, Břehová 7, Prague, 110 00, Czech Republic
}

\maketitle

%% file: src/abstract.tex
\begin{abstract}
    Last decade has seen the emergence of numerous methods for learning on graphs, particularly Graph Neural Networks (GNNs). These methods, however, are often not directly applicable to more complex structures like bipartite graphs (equivalent to hypergraphs), which represent interactions among two entity types (e.g.\ a user liking a movie). This paper proposes \methodlong{} (\method{}), a non-parametric simple and scalable method that natively operates on bipartite graphs (hypergraphs) and can be implemented with just a few lines of code. After defining \method{}, we demonstrate its relationship with well-established methods like label propagation, Naive Bayes, and Hypergraph Convolutional Networks. We evaluate \method{} against several reference methods on real-world datasets from multiple domains, focusing on retrieval and classification tasks. Our results show that \method{} offers competitive performance while maintaining low computational complexity, making it an ideal first choice as a baseline for hypergraph node classification and retrieval. Moreover, despite operating on hypergraphs, CSP achieves good results in tasks typically not associated with hypergraphs, such as natural language processing.

    \keywords{%
        Hypergraph representation learning \and
        Hypergraph convolution \and
        Label Propagation \and
        Model complexity \and
        Naive Bayes
    }
\end{abstract}

%% file: src/introduction.tex
\section{Introduction}
In the modern world, an overwhelming amount of data has an internal structure, oftentimes forming complex networks that can be represented as graphs. Efficiently mining information from this data is crucial for a wide range of applications spanning various domains such as social networks, biology, physics or cybersecurity. Graph Neural Networks (GNNs) have emerged as the dominant tool for handling such data due to their ability to leverage the graph structure for predictive and analytical tasks. However, despite their success, GNNs come with notable challenges, including high computational complexity during training, numerous hyperparameters that require fine-tuning, lack of straightforward interpretability, and the necessity of dedicated computational infrastructure such as GPUs.

Given these limitations, baseline algorithms play a vital role as complementary tools to GNNs. These baselines, often much less complex, provide an efficient way of generating preliminary results. In many cases, these simpler methods are even sufficiently effective to be used as-is for the problem at hand. 

Section \ref{sec:problem-statement} introduces the problem being solved in this work and Section \ref{sec:related-work} provides an overview of other related works. Section \ref{sec:our-method} is the most substantive part of this paper, introducing the \method{} method (Section \ref{sec:cop}). In Section \ref{sec:label-prop}, we show \method{} to be a straightforward extension of the well-known label propagation method to hypergraphs. Also provided is a comparison of \method{} to other methods, such as hypergraph convolutional networks (Section \ref{sec:HGCNvsCP}) and the naive Bayes classifier (Section \ref{sec:naive-bayes}). Finally, Section \ref{sec:experimental-evaluation} provides an experimental evaluation and comparison of \method{} and several alternative methods.

%% file: src/problem-statement.tex
\section{Problem Statement}\label{sec:problem-statement}

Assume that we have structured data with relationships that can be translated into a hypergraph or a bipartite graph. These structures can represent various scenarios such as users rating movies, users accessing web domains, emails containing attachments, authors writing papers, papers being co-cited, and tokens being contained in texts. Some data might also come with constraints such as large volume or privacy restrictions, like those found in emails. These structures inherently provide a relationship between entities, enabling many applications to leverage these relationships, such as movie recommendations, mining malicious web content (like emails or domains), or text classification.

Our goal is to find a flexible method that can be applied to tasks involving structured data. This flexibility is sought in terms of performance, numerical complexity, and adaptability. We aim to develop a method that can efficiently handle and extract meaningful information from these complex relationships. Whether we are looking to extract interesting entities, which aligns with a retrieval scenario, or view the task as a simple classification problem, the method should be versatile enough to adapt to these needs.

We can formalize this problem using either a bipartite graph or a hypergraph. By representing the data in these graph structures, we can better understand and utilize the inherent relationships between entities. This formalization allows for the application of graph-based algorithms, which can improve the effectiveness of tasks like classification and retrieval.


\subsection{Notations and Definitions} \label{Sec:Notation}
We consider a finite set of items of interest $\U=\{\uu_1, \dots, \uu_n\}$, referred to as \name{nodes}. A family of subsets of $\U$ denoted by $\V=\{\vv_1, \dots, \vv_m\} \subseteq 2^\U$ is referred to as \name{hyperedges}. The nodes and hyperedges togerher form a \name{hypergraph} $\HG=(\U, \V)$. The structure of the hypergraph can also be described by an \name{incidence matrix} $\HH \in \{0,1\}^{n\times m}$, where $\HH_{i, j} = 1$ if $\uu_i \in \vv_j$, and $0$ otherwise. Every hypergraph can alternatively be described by a bipartite \name{incidence graph}, also called the Levi graph \cite{levi_finite_1942}. This bipartite graph $\BG=(\U \cup \V, \E)$ has as its two partitions the nodes and hyperedges of $\HG$, and its edges represent a node in $\U$ belonging to an edge in $\V$, formally $\E=\left\{\left(\uu_i, \vv_j\right) \in \U \times \V \middle| \uu_i \in \vv_j \right\}$. 

The degree \( \vdeg \left( \uu \right) \) of node \( \uu \) is defined as the number of edges that contain the node, that is \( \vdeg \left( \uu \right) = \left\lvert \left\{ \vv_i \in \V \middle| \uu \in \vv_i \right\} \right\rvert \). Similarly, the degree \( \edeg \left( \vv \right) \) of the edge \( \vv \) is defined as the number of nodes it contains, i.e. \( \edeg \left( \vv \right) = \left\lvert \vv \right\rvert \). We also establish a diagonal node-degree matrix \( \D \in \mathfield{N}^{n \times n} \) with \( (\D)_{i,i} = \vdeg \left( \uu_i \right) \) and \( (\D)_{i,j} = 0 \) for \( i \neq j \). Analogously, the hyperedge-degree matrix is a diagonal matrix \( \B \in \mathfield{N}^{m \times m} \) with \( (\B)_{i,i} = \edeg \left( \vv_i \right) \) and \( (\B)_{i,j} = 0 \) for \( i \neq j \).

We consider for each node in the hypergraph some kind of signal that is to be propagated through the hyperedges. Let the signal be a \( d \)-dimensional vector \( \mathvec{x}_i \) for each node, giving for the whole hypergraph a matrix \( \X \in \mathfield{R}^{n \times d} \). In the following parts of this paper, we will explore several ways of defining such a signal, with an overview provided in Section \ref{sec:cop-matrix-realizations}.

Within this work, we are interested in two transductive tasks on hypergraphs: classification on $\U$ and retrieval of positive nodes from $\U$. Both tasks assume a training set of nodes $\Tr \subset \U$ where the labels of nodes are known. In case of classification, the goal is to predict the label for all nodes in $\U$. The retrieval task aims to sort the nodes in the testing set $\U \setminus \Tr$ such that the number of positive nodes in top $K$ positions is maximized.

%% file: src/related-work.tex
\section{Related work}\label{sec:related-work}
Mining information from structured (graph-like) data is one of the central problems in machine learning. The most straightforward way to handle this is to translate the structure into features and apply traditional machine learning techniques, such as logistic regression, random forests, and naive Bayes, to these features. Naive Bayes \cite{ng2001discriminative}, in particular, provides a bridge to a second large family of learning methods on graphs: Bayesian methods, where the graph forms a structure for modelling random variables. A critical problem associated with Bayesian methods is inference, which is often intractable. The sum product algorithm  \cite{kschischang2001factor} combats the tractability of a system by its decomposition to a product of local functions. One instance of this framework is probabilistic threat propagation \cite{carter2014probabilistic}, which is used for spreading information about malicious actors through the network.

The translation of structured data into features is also a non-trivial problem. While some methods can handle sparse, high-dimensional feature vectors, the majority cannot. Several methods are suited for finding low-dimensional representations of structured data. For example, non-negative matrix factorization \cite{lee2000algorithms} (NMF) decomposes a large sparse matrix into the product of two low-dimensional matrices. Node2vec \cite{grover2016node2vec} applies contrastive learning to find node representations where neighboring nodes are close in feature space. Spectral positional encodings \cite{dwivedi_benchmarking_2023} provide representations given by eigenvectors of the Laplacian matrix, and distance encodings \cite{li2020distance,beaini_directional_2021} offer global information about where a node is located in the graph. However, except for NMF, these methods are suited for graphs and cannot be directly applied to hypergraphs.

There are many papers on learning algorithms for graphs, such as GraphSAGE \cite{hamilton2017inductive}, graph convolutional networks \cite{kipf2016semi}, and graph attention networks \cite{velivckovic2017graph}. Nevertheless, their application to hypergraphs is not straightforward. The origins of learning transductive tasks stretch back to the seminal work \cite{zhou_learning_2006}. More recently, Hypergraph neural networks \cite{feng_hypergraph_2019}, Dynamic HGNNs \cite{jiang_dynamic_2019}, and HyperGCN \cite{yadati_hypergcn_2019} build upon the convolutional learning schema introduced in \cite{kipf_semi-supervised_2017} while works such as \cite{bai_hypergraph_2021} aim to bring both convolutional as well as attention to the context of hypergraphs. The proposed method can also be viewed as an extension of label propagation \cite{zhu_learning_2003,huang_combining_2020} or feature propagation \cite{rossi_unreasonable_2022}. While there do exists algorithm for label propagation in hypergraphs \cite{henne_label_2015,lee_villain_2024}, the proposed method aims to be comparatively simpler to understand, implement and calculate. The proposed method is an extension of our previous work in the specific domain of computer network security \cite{prochazka_cross-domain_2023}.

%% file: src/our-method.tex
\section{\methodlong{}}\label{sec:our-method}

We present \methodlong{} (\method{}), a method for signal propagation on hypergraphs. In the following subsections, \method{} is first introduced in the general setting, followed by a comparison to established approaches and a discussion of possible variants inspired by them. Finally, applications of \method{} to different kinds of signals in hypergraph tasks are discussed.

\subsection{Method overview}\label{sec:cop}
The proposed algorithm propagates a node signal $\X$ (see Section \ref{sec:cop-matrix-realizations} for a discussion of possible signal types) through the hypergraph $\HG$. The basic version of \method{} consists in a simple averaging of \( \X \) across the hyperedges and nodes of the graph. This averaging can be repeated to obtain smoother final representations, resulting in a multi-step \method{} process generating a sequence of representations \( \X^{(l)} \), where \( \X^{(0)} = \X \).

In each step, the representation $\X^{(l)}$ of the nodes is first propagated to the hyperedges to obtain their representations 
\begin{equation}\label{eq:edge_score}
\mathvec{r}_j^{(l)}=\frac{1}{\edeg \left( \vv_j \right)}\sum_{\substack{i \\ \uu_i\in \vv_j}} \x_i^{(l)} 
\end{equation}
that is the average of the representation of the individual nodes contained in the hyperedge. In the second step, this hyperedge representation is propagated again into nodes:
\begin{equation}\label{eq:nodes_score}
\x_k^{(l+1)}=\frac{1}{\vdeg \left( \uu_k \right)}\sum_{\substack{j \\ \uu_k \in \vv_j}} \mathvec{r}_j^{(l)}.
\end{equation}

The steps \ref{eq:edge_score} and \ref{eq:nodes_score} constitute the proposed \methodlong{} algorithm, which can be summarily written as
\begin{equation}\label{eq:CP}
    \x_k^{(l+1)} = \frac{1}{\vdeg \left( \uu_k \right)}\sum_{\substack{j \\ \uu_k \in \vv_j}} \frac{1}{\edeg \left( \vv_j \right)}\sum_{\substack{i \\ \uu_i \in \vv_j}} \x_i^{(l)}.
\end{equation}

Using notation established in Section \ref{Sec:Notation}, Equation \ref{eq:CP} can be rewritten into the matrix form
\begin{equation}\label{eq:CP_Matrix_Form}
    \X^{(l+1)} = \D^{-1} \HH \B^{-1} \HH^T \X^{(l)}.
\end{equation}

Equation \ref{eq:CP_Matrix_Form} describes a basic variant of the proposed algorithm. In Section \ref{sec:CoP-variants}, various modifications of \method{} are discussed.

\subsection{On Efficient Implementation of \methodlong{}}
While Equation \ref{eq:CP_Matrix_Form} shows an efficient way of mathematically expressing the \method{} algorithm, the algorithm itself is also efficient when it comes to its implementation and computational complexity. Algorithm \ref{alg:CP-SQL} shows an implementation of \method{} in a single SQL query and Algorithm \ref{alg:CP-python} shows a simple implementation in Python using the Pandas \cite{mckinney_data_2010} library. These implementations essentially materialize Equations \ref{eq:edge_score} and \ref{eq:nodes_score}.

\begin{algorithm}
    \caption{An SQL implementation of \method{} (\ref{eq:CP}), where graph is a table with columns nodeId, edgeId and property. While nodeId and edgeId desribe the hypergraph, property defines the task.}
    \label{alg:CP-SQL}
    \begin{lstlisting}[language=SQL,numbers=left]
SELECT nodeId, AVG(edgeProperty) AS updatedProperty
FROM (
    SELECT nodeId, AVG(property) OVER (PARTITION BY edgeId) AS edgeProperty
    FROM graph
)
GROUP BY nodeId;
    \end{lstlisting}
\end{algorithm}

\begin{algorithm}
    \caption{A Pandas implementation of \method{} (\ref{eq:CP}), where df is a DataFrame with columns nodeId, edgeId and property. While nodeId and edgeId desribe the hypergraph, property defines the task.}
    \label{alg:CP-python}
    \begin{lstlisting}[language=python,numbers=left]
import pandas as pd
def CSP_layer(df: pd.DataFrame) -> pd.DataFrame:
    return df.assign(edgeProperty=df.groupby('edgeId')['property'].transform('mean')) \
        .groupby('nodeId')['edgeProperty'].mean() \
        .reset_index().rename(columns={'edgeProperty': 'updatedProperty'})
    \end{lstlisting}
\end{algorithm}

\subsection{Application of \method{} to different signals in hypergraphs}
\label{sec:cop-matrix-realizations}

The construction of \methodlong{} in Section \ref{sec:cop} was a general one, assuming a signal matrix \( \X \in \mathfield{R}^{n \times d} \). In practice, one can use \method{} to propagate various kinds of \enquote{signals} in the hypergraph. Namely, the matrix \( \X \) may represent actual features as provided in the underlying graph dataset. This setting leads to a method similar to feature propagation \cite{rossi_unreasonable_2022} or hypergraph convolution \cite{feng_hypergraph_2019}. Such an approach is elaborated further in Section \ref{sec:HGCNvsCP}. Alternatively, an approach based on label propagation \cite{zhu_learning_2003} may be obtained by taking as \( \X \) a version of the label matrix \( \Y \) masked by the training set, a setting described in Section \ref{sec:label-prop}. Finally, the hyperedges of \( \HG \) may be used as the signal, essentially representing indicators of node similarity based on their presence in a common hyperedge. Such an approach corresponds to setting \( \X = \HH \) and is described in Section \ref{sec:naive-bayes}.

\subsection{Comparison with Hypergraph Convolution} \label{sec:HGCNvsCP}
A single layer of the Hyper-Conv hypergraph neural network by \cite{bai_hypergraph_2021} is defined as 
\begin{equation}
    \X^{(l+1)} = \sigma(\D^{-1} \HH \mathmat{W} \B^{-1} \HH^T \X^{(l)} \boldsymbol{\Theta}) \label{eq:HGCN},
\end{equation}
where $\mathmat{W}$ and $ \boldsymbol{\Theta} $ are weight matrices that need to be optimized.

Comparing Equations \ref{eq:CP_Matrix_Form} and \ref{eq:HGCN}, it can be seen that \method{} is a simplified special case of Hyper-Conv with the matrices \( \mathmat{W} \) and \( \boldsymbol{\Theta} \) realized as non-learnable identity matrics. As the proposed method runs only the \enquote{forward pass} of Hyper-Conv, we do not use the non-linearity $\sigma$ in the basic variant of \method{}. 

\subsection{Comparison with Label Propagation}\label{sec:label-prop}
The label propagation algorithm as intruduced in \cite{zhu_learning_2003} is expressed for an ordinary graph (with edges connecting exactly 2 nodes) as 
\begin{equation} \label{eq:LP}
    \Y^{(l+1)} = \alpha \D^{-1} \mathmat{A} \Y^{(l)} + \left( 1 - \alpha \right) \Y^{(l)},
\end{equation}
where $\D$ denotes the diagonal matrix of degrees of a graph and $\mathmat{A}$ stands for its adjacency matrix.

To compare Label propagation with \method{}, let us first express the value of \( \HH \B^{-1} \HH^T \) as
\begin{equation}\label{eq:hypergraph-adjacency-matrix}
    \left( \HH \B^{-1} \HH^T \right)_{i, j} = \sum_k \frac{1}{\edeg \left( \vv_k \right)} \HH_{i, k} \HH_{j, k},
\end{equation}
which represents for each pair of nodes the number of hyperedges connecting them, normalized by their degrees. Specifically, for an ordinary graph, this becomes
\begin{equation}
    \HH \B^{-1} \HH^T = \frac{1}{2} \left( \mathmat{A} + \D \right).
\end{equation}
With this simplification for ordinary graphs, equation \ref{eq:CP_Matrix_Form} becomes
\begin{equation}
    \X^{(l+1)} = \frac{1}{2} \D^{-1} \mathmat{A} \X^{(l)} + \frac{1}{2} \X^{(l)}.
\end{equation}
which is equivalent to equation \ref{eq:LP} with \( \X = \Y \) (or a masked version thereof) and \( \alpha = \frac{1}{2} \). \method{} is in this instance therefore a generalization of label propagation with this particular value of \( \alpha \) to hypergraphs (for generalization with arbitrary values of \( \alpha \), see Section \ref{sec:CoP-variants}). There is, however, another compelling reason to use \method{} over Label propagation as presented in equation \ref{eq:LP}. The matrix multiplication $\HH \B^{-1} \HH^T$ does not preserve the sparsity of $\HH$, which is typical for large datasets. Therefore the proposed implementation can be significantly more efficient than Equation \ref{eq:LP}, despite them being mathematically equivalent.

\subsection{Comparison with the naive Bayes classifier}\label{sec:naive-bayes}
The naive Bayes classifier is a well know classification method which calculates posterior probability on $\y$ with a feature vector $\xi$ as $p(\y|\xi)$ using Bayes rule as $p(\y|\xi)=\frac{p(\xi|\y)p(\y)}{p(\xi)}$ with the \enquote{naive} assumption that $p(\xi|\y)=\prod_i p(\xi_i|\y)$, where  $p(\xi_i|\y)$ is estimated on training set for all pairs of features and labels. 

In the case of a multinomial Naive Bayes, the term $p(\xi_i|\Y)$ is given by the empirical distribution observed on the training set. In the \method{} setup, we therefore consider the hyperedges of \( \HG \) to represent the features \( \xi \), that is \( \X = \HH \). In the binary case, the empirical distribution $p(\xi_i|\Y)$ is equal to averaging the node signal across each hyperedge as described in Equation \ref{eq:edge_score}. The aggregation (product) then nicely demonstrates the difference between the filtering (convolutional) approach of \method{} and the Bayesian (Maximum a posteriori) approach.

\subsection{\methodlong{} Variants} \label{sec:CoP-variants}

The comparison with the methods presented in Sections \ref{sec:HGCNvsCP}, \ref{sec:label-prop} and \ref{sec:naive-bayes} naturally suggests several alternative variants and generalizations of the basic \method{} scheme.

\subsubsection{Alternative normalizations of the adjacency matrix}

In graph neural networks, the adjacency matrix \( \mathmat{A} \) is normalized by multiplying it with the inverse of the node degree matrix \( \D \). While the DeepWalk algorithm \cite{perozzi_deepwalk_2014} corresponds to the row-wise normalization \( \D^{-1} \mathmat{A} \), newer methods also consider the column-wise normalization \( \mathmat{A} \D^{-1} \) and most predominantly the symmetric normalization \( \D^{-1/2} \mathmat{A} \D^{-1/2} \) introduced in \cite{kipf_semi-supervised_2017}. Because the matrix \( \HH \B^{-1} \HH^T \) plays in \method{} a role similar to the adjacency matrix \( \mathmat{A} \) in GCN (see Equation \ref{eq:hypergraph-adjacency-matrix}), we can also consider the alternative column-wise normalized version of \method{}:
\begin{equation}\label{eq:CoP-column}
    \X^{(l+1)} = \HH \B^{-1} \HH^T \D^{-1} \X^{(l)}
\end{equation}
and the symmetrically normalized version
\begin{equation}\label{eq:CoP-symmetrical}
    \X^{(l+1)} = \D^{-1/2} \HH \B^{-1} \HH^T \D^{-1/2} \X^{(l)}.
\end{equation}

\subsubsection{Generalization of label propagation with general values of $\alpha$}

While Section \ref{sec:label-prop} shows that \method{} is generalization of label propagation with \( \alpha = \frac{1}{2} \) to hypergraphs, a more general variant of \method{} allowing for \( \alpha \in \left( 0, 1 \right) \) can be defined as
\begin{equation}\label{eq:CoP-general-label-prop}
    \X^{(l+1)} = 2 \alpha \D^{-1} \HH \B^{-1} \HH^T \X^{(l)} + \left( 1 - 2 \alpha \right) \X^{(l)}.
\end{equation}
This version is a full generalization of label propagation as described in Equation \ref{eq:LP} to hypergraphs (equivalence can be proved by applying Equation \ref{eq:hypergraph-adjacency-matrix}).


All of the above variants of \method{} can be implemented in a straightforward way by modifying Algorithms \ref{alg:CP-SQL} and \ref{alg:CP-python}, without requiring full matrix multiplication.

%% file: src/experimental-evaluation.tex
\section{Experimental evaluation}\label{sec:experimental-evaluation}
The goal of our experiments is to compare performance and execution time of the proposed \methodlong{} (\method{}) method with several well established baseline methods as well as with a simple Hypergraph Neural Network (HGCN) on a variety of real-world datasets from multiple domains. In this section, we introduce the considered datasets, reference methods, and tasks used for evaluation. Our aim is to validate the comparable performance of the proposed method while highlighting its low execution time. While we discuss the various variants of the proposed method in Section \ref{sec:CoP-variants}, a comprehensive evaluation of these variants will be conducted in a future work.

\subsection{Datasets}

\begin{table}[]
    \caption{Overview of datasets with their basic characteristics. \( \Sigma_\V = \sum_{\vv \in \V} \edeg \left( \vv \right) \) or equivalently the number of non-zero elements in the incidence matrix $\HH$. Isolated nodes are nodes that are not connected by any hyperedge. Percentage of isolated nodes is their fraction related to the overall number of nodes.}
    \label{tab:dataset_overview}
    \centering
        \begin{tabular}{lrrrrrrrr}
        \toprule
        \textbf{Dataset} & \textbf{CiteSeer} & \textbf{Cora-CA}  & \textbf{Cora-CC}  & \textbf{DBLP} & \textbf{PubMed}   & \textbf{Corona} & \textbf{movie-RA} & \textbf{movie-TA} \\
        \midrule
        \textbf{Nodes}  & 3312 & 2708 & 2708 & 41302 & 19717 & 44955 & 62423 & 62423 \\
        \textbf{Isolated nodes} & 1854 & 320 & 1274 & 0 & 15877 & 0 & 3376 & 17172 \\
        \textbf{Isolated node ratio} & 56\% & 12\% & 47\% & 0\% & 81\% & 0\% & 5\% & 28\% \\
        \textbf{Hyperedges}  & 1079 & 1072 & 1579 & 22363 & 7963 & 998 & 162541 & 14592\\
        \( \boldsymbol{\Sigma}_\V \)  & 3453 & 4585 & 4786 & 99561 & 34629 & 3455918 & 25000095 & 1093360 \\
        \textbf{Average \( \vdeg \left( \uu \right) \)} & 2.37 & 1.92 & 3.34 & 2.41 & 9.02 & 76.88 & 423.39 & 24.16 \\
        \textbf{Average \( \edeg \left( \vv \right) \)} & 3.20 & 4.28 & 3.03 & 4.45 & 4.35 & 3463 & 153.8 & 74.9 \\
        \textbf{Classes} & 6 & 7 & 7 & 6 & 3 & 5 & 20 & 20 \\
        \bottomrule
        \end{tabular}
\end{table}

We evaluated our methods on datasets from three distinct areas:
\begin{itemize}
    \item \textbf{Citation datasets}: This category includes the PubMed, Cora, and DBLP datasets in their hypergraph variants, as introduced in \cite{chien_you_2021}. In these datasets, the hyperedges may be defined in two distinct ways - either by a common author, or by co-citation, yielding two variants of the dataset. Each publication is labeled based on its topic.
    \item \textbf{Natural Language Processing (NLP)}: We used the Coronavirus tweets NLP dataset \cite{miglani_coronavirus_2020}, which contains Twitter posts about COVID-19. The tweets were tokenized, with each tweet representing a node, labeled by its sentiment. Hyperedges are formed by tokens appearing in multiple tweets. The tokens are trained on the whole corpus with given number of tokens\footnote{The number differs to the number of hyperedges from Table \ref{tab:dataset_overview}, due to special reserved tokens that are not used in corpus.} (1000) using  Sentencepiece algorithm \cite{kudo_sentencepiece_2018}. Note that we can control the graph size with (number of hyperedges) by the parameter of tokenization. 
    \item \textbf{Recommender Systems}: We used the MovieLens 25M dataset \cite{harper_movielens_2015}, which contains 25 million movie ratings and one million tag applications. The nodes are individual movies labeled by their genres (allowing for multiple labels for one node). The hyperedges correspond to users and may again be defined in two ways, giving us two versions of the dataset: Either by a user rating multiple movies, or by a user assigning tags to multiple movies.
\end{itemize}

An overview of datasets including their basic characteristics is shown in Table  \ref{tab:dataset_overview}. Due to the interpolative nature of \method{}, each multiclass dataset is transformed into a series of binary datasets, where each class is treated as the positive class, and all other classes are treated as negative. The results are then averaged over all such obtained binary datasets.

We consider only structural information that is available in the hypergraph. No other information (such as node or hyperedge features) is included. 

\subsection{Tasks}

We address two primary tasks in our experiments: transductive node classification and retrieval.

\subsubsection{Classification Task}

\begin{itemize}
    \item \textbf{Aim}: Prediction of binary labels on the testing set
    \item \textbf{Method}: We use leave-one-out cross-validation with 10 folds, where nodes are randomly assigned to folds. One fold is hidden for testing, and the method is trained on the remaining nine folds. For each dataset, we generate test predictions for each node (when it is in the testing set). For each class, we calculate the ROC-AUC and average these scores. This average is reported as the classification score for each method on the given dataset.
\end{itemize}

\subsubsection{Retrieval Task}

\begin{itemize}
    \item \textbf{Aim}: Ranking of nodes in the testing set maximizing the precision in the top positions
    \item \textbf{Method}: Folds are assigned in the same way as in the classification task, with one fold being used for training and the other 9 used as the testing dataset. The training set consists of all nodes in the graph, with the positive nodes in the training fold labeled as positive and all other nodes labeled as unknown. The model is trained on this training set. For models that require negative training examples, we randomly sample a set of the same size as the testing set and consider these labels as negative. Although this introduces some label noise, we assume that the negative class is dominant, making the noise acceptable. The model then ranks the nodes from the testing folds, and we evaluate precision at the top 100 positions (P@100). This evaluation is performed for each fold and class, and the average P@100 over folds and classes is reported as the retrieval score for each method on the given dataset.
\end{itemize}

\subsection{Evaluated Methods}

We compare the following methods in our experiments. The variants of the \method{} mentioned in Section \ref{sec:CoP-variants} are not reported in this work. In future, we would like also to compare multiple choices of feature representation for reference methods on top of NMF. Nevertheless as our goal is just to show a competitive performance of \method{}, we believe that it is sufficient to compare it with baseline methods without their fine-tuning.

\begin{itemize}
    \item \textbf{The proposed \method{} method}: Evaluated with 1, 2, and 3 layers, where we consider binary training labels as $\X^{0}$. After application of a given number of \method{} layers (see Equation \ref{eq:CP_Matrix_Form}), the yielded (score) vector $\X^{l}, l\in\{1,2,3\}$ is used for both retrieval (top-100 scored test nodes) and for binary classification (with a given threshold on score).
    \item \textbf{Multinomial Naive Bayes}: Operates on one-hot feature vectors derived from hyperedges.
    \item \textbf{Random Forest, Logistic Regression, and HGCN}: These methods operate on feature vectors obtained from non-negative matrix factorization (NMF) of the incidence matrix\cite{lee2000algorithms}\footnote{As an alternative to the NMF, we evaluated also representation generated by Laplacian positional encoding \cite{dwivedi_benchmarking_2023}. As the results were worse compared to NMF, we decided to not include them in the results.} $\HH$, with 10 iterations and a dimension of 60.
    \item \textbf{Method Settings}:
    \begin{itemize}
        \item Random Forest, Logistic Regression, and Naive Bayes are used with their default settings.
        \item A single layer HGCN implements Equation (\ref{eq:HGCN}) with an output layer of dimension 2 and sigmoid non-linearity. We use logistic loss and train all datasets for 15,000 epochs using the Adam optimizer with default settings.
    \end{itemize}
    \item \textbf{Random Baseline}: Included for comparison.
\end{itemize}

The results of these methods are evaluated and compared based on their performance on the classification and retrieval tasks across the datasets.

\subsection{Classification Result} 

Table \ref{tab:classification-results} lists the ROC-AUC for all methods on all datasets. Due to the numerical intensity of HGCN, only 5 folds were evaluated, and for the Movies dataset, only 4 out of 20 classes were considered. Prediction on isolated nodes was nearly random as only structural information was used, likely contributing to the relatively weak performance of all methods on the PubMed dataset. Since CSP handles only binary labels, reference methods were translated to a one-vs-other scenario, even though they can handle multi-class classification directly. Feature extraction using Non-negative Matrix Factorization (NMF) was not fine-tuned for each dataset, potentially impacting the performance of NMF-based baselines. Naive Bayes emerged as the strongest baseline, as it does not require any feature preprocessing and works directly with the one-hot encoded incidence matrix. CSP was evaluated in three variants based on the number of layers, with the best choice varying by dataset. On the largest datasets (Corona and Movies), the best variant of CSP achieved performance comparable to the strongest competing baseline. Overall, CSP demonstrated comparable results with reference baselines. In larger datasets, where parameter tuning of baselines is more challenging, CSP proved to be one of the best-performing methods. Overall, CSP with fewer layers fared comparatively better than a version with multiple layers. We attribute this at first glance counter-intuitive result to the fact that the training set is fairly dense in the graph, which ensures sufficient information for all nodes even with fewer layers, while at the same time multiple layers may contribute to oversmoothing of the signal. CSP's simplicity and parameter-free nature confirm its suitability as a first-choice baseline method for classification tasks. 

\begin{table}
    \caption{The ROC-AUC for the classification task, averaged over all classes and all folds. The best method for each dataset is denoted by bold text, with methods within \( 0.05 \) underlined.}
    \label{tab:classification-results}
    \centering
    \begin{tabular}{lrrrrrrrr}
        \toprule
        \textbf{Method} & \textbf{CiteSeer} & \textbf{Cora-CA} & \textbf{Cora-CC} & \textbf{DBLP} & \textbf{PubMed} & \textbf{Corona} & \textbf{movie-RA} & \textbf{movie-TA} \\
        \midrule
        \textbf{\method{} 1-layer} & \underline{0.646} & \underline{0.882} & 0.716 & \underline{0.968} & 0.537 & \textbf{0.704} & \underline{0.789} & \underline{0.717} \\
        \textbf{\method{} 2-layer} & 0.630 & \underline{0.872} & 0.686 & \underline{0.972} & 0.518 & 0.618 & 0.700 & \underline{0.697} \\       
        \textbf{\method{} 3-layer} & 0.613 & 0.862 & 0.655 & \underline{0.972} & 0.516 & 0.580 & 0.640 & 0.673 \\
        \textbf{Naive Bayes} & \textbf{0.686} & \textbf{0.913} & \underline{0.775} & \textbf{0.974} & \textbf{0.633} & \textbf{0.704} & \underline{0.753} & 0.557 \\   
        \textbf{HGCN-NMF} & \underline{0.659} & 0.832 & \textbf{0.786} & 0.775 & \underline{0.624} & 0.622 & \underline{0.794} & \textbf{0.724} \\
        \textbf{LR-NMF} & 0.604 & 0.794 & 0.703 & 0.705 & 0.556 & 0.647 & \underline{0.754} & \underline{0.675} \\
        \textbf{RF-NMF} & \underline{0.667} & \underline{0.897} & \underline{0.772} & 0.905 & \underline{0.623} & 0.617 & \textbf{0.797} & \underline{0.691} \\
        \textbf{Random} & 0.499 & 0.505 & 0.489 & 0.501 & 0.502 & 0.503 & 0.499 & 0.487 \\
        \bottomrule
    \end{tabular}
\end{table}

\subsection{Retrieval Results}

Table \ref{tab:retrieval-results} lists the P@100 for all methods on all datasets. 
 The evaluation of HGCN and the Movies dataset in the retrieval task is restricted similarly to the classification task. Isolated nodes no longer cause a performance drop as long as there are sufficient number of non-isolated nodes in each class. Naive Bayes' performance is not as superior in this scenario as in classification task since the training set contains only positive nodes, preventing it from leveraging prior distribution knowledge about the target class. CSP, which does not use prior knowledge about the target class distribution, works very well on small datasets with lower average degree of the nodes and edges. In case of datasets with higher average degree of nodes (Movies), CSP does not extract the structural information as well as NMF and therefore the methods utilizing the features from NMF (mainly logistic regression) work much better excepting HGCN, which suffers from over-smoothing. In summary,  \method{} achieves superior performance for 4 of 8 datasets on retrieval task. 

\begin{table}
    \caption{The P@100 for the retrieval task, averaged over all classes and all folds. The best method for each dataset is denoted by bold text, with methods within \( 0.05 \) underlined.}
    \label{tab:retrieval-results}
    \centering
    \begin{tabular}{lrrrrrrrr}
        \toprule
        \textbf{Method} & \textbf{CiteSeer} & \textbf{Cora-CA} & \textbf{Cora-CC} & \textbf{DBLP} & \textbf{PubMed} & \textbf{Corona} &\textbf{movie-RA} & \textbf{movie-TA} \\
        \midrule
        \textbf{\method{} 1-layer} & 0.494 & \underline{0.703} & 0.530 & 0.869 & 0.798 & \textbf{0.530} & 0.334 & 0.156 \\
        \textbf{\method{} 2-layer} & \underline{0.558} & \underline{0.718} & \underline{0.681} & 0.865 & \underline{0.826} & 0.440 & 0.336 & 0.186 \\       
        \textbf{\method{} 3-layer} & \textbf{0.568} & \textbf{0.721} & \textbf{0.707} & 0.869 & \underline{0.850} & 0.332 & 0.238 & 0.186 \\       
        \textbf{Naive Bayes} & 0.471 & \underline{0.686} & 0.491 & \textbf{0.951} & \underline{0.860} & 0.446 & 0.216 & 0.153 \\
        \textbf{HGCN-NMF} & 0.482 & \underline{0.671} & 0.607 & 0.794 & \textbf{0.871} & 0.392 & 0.257 & 0.148 \\
        \textbf{LR-NMF} & 0.329 & 0.603 & 0.372 & 0.602 & 0.735 & 0.397 & \textbf{0.580} & \textbf{0.356} \\
        \textbf{RF-NMF} & 0.303 & 0.474 & 0.482 & 0.843 & 0.794 & 0.381 & 0.470 & 0.131 \\
        \textbf{Random} & 0.153 & 0.132 & 0.129 & 0.155 & 0.308 & 0.180 & 0.040 & 0.055 \\
        \bottomrule
    \end{tabular}
\end{table}

\subsection{Complexity Evaluation}

\begin{table}
    \caption{Asymptotic complexity of the compared algorithms. For algorithms with multiple layers, only one layer is considered. \( n \) is the number of nodes, \( m \) is the number of hyperedges \( d \) is the signal dimensionality, \( T \) is the number of epochs, \( h \) is the number of hidden units, \( \Sigma_\U \) is the sum of node degrees and \( \Sigma_\V \) is the sum of hyperedge degrees.}
    \label{tab:asymptotic-complexity}
    \centering
    \begin{tabular}{lccccc}
        \toprule
        \textbf{\method{}} & \textbf{Naive Bayes} & \textbf{HGCN} & \textbf{LR} & \textbf{RF} \\
        \midrule
        \( \mathcal{O} \left( d \left( \Sigma_\U + \Sigma_\V \right) \right) \) & \( \mathcal{O} \left( n m \left\lvert \Tr \right\rvert \right) \) \cite{fleizach_naive_1998} & \( \mathcal{O} \left( T \Sigma_\V h d \right) \) \cite{yadati_hypergcn_2019} & \( \mathcal{O} \left( d \left\lvert \Tr \right\rvert \right) \) \cite{nocedal_numerical_1999} & \( \mathcal{O} \left( d \log{d} \left\lvert \Tr \right\rvert \right) \) \cite{biau_analysis_2012} \\
        \bottomrule
    \end{tabular}
\end{table}

\begin{table}
    \caption{The execution time of a single retrieval task in microseconds, averaged over all classes and all folds. The non-negative matrix factorization was excluded from the execution time.}
    \label{tab:execution-time}
    \centering
    \begin{tabular}{lrrrrrrrr}
        \toprule
        \textbf{Method} & \textbf{CiteSeer} & \textbf{Cora-CA} & \textbf{Cora-CC} & \textbf{DBLP} & \textbf{PubMed} & \textbf{Corona} & \textbf{movie-RA} & \textbf{movie-TA} \\
        \midrule
        \textbf{\method{} 1-layer} & \textbf{1.35} & \textbf{1.23} & \textbf{1.24} & \textbf{3.51} & \textbf{4.01} & \textbf{22.83} & 170.63 & \textbf{12.07} \\             
        \textbf{\method{} 2-layer} & 2.41 & 2.25 & 2.24 & 7.46 & 7.35 & 47.39 & 349.67 & 25.88 \\
        \textbf{\method{} 3-layer} & 3.31 & 3.09 & 3.08 & 9.65 & 9.1 & 70.98 & 506.1 & 35.75 \\
        \textbf{Naive Bayes} & 2.59 & 2.73 & 2.54 & 11.16 & 5.35 & 89.3 & 1 051.53 & 35.61 \\
        \textbf{HGCN-NMF} & 23 714 & 23 690 & 23 709 & 29 123 & 23 879 & 112 314 & 620 717 & 70 778 \\
        \textbf{LR-NMF} & 44.45 & 51.59 & 49.54 & 64.96 & 44.69 & 56 & \textbf{61.82} & 74.07 \\
        \textbf{RF-NMF} & 140.76 & 143.85 & 134.52 & 1 148.83 & 323.6 & 1 608.58 & 697.85 & 716.68 \\
        \bottomrule
    \end{tabular}
\end{table}

To evaluate the complexity of the studied methods, two approaches were taken -- theoretical analysis of asymptotic complexity, and direct measurement of wall clock time to compute the methods on the available datasets. The asymptotic computational complexities of the used methods are listed in Table \ref{tab:asymptotic-complexity}. Note that the non-negative matrix factorization is excluded from the table and has on its own a complexity of \( \mathcal{O} \left( n^4 m^4 \right) \), dominating the methods themselves. The execution time for individual methods is presented in Table \ref{tab:execution-time}. We evaluated the methods using standard implementations that would be widely used by practitioners. In particular, we used the Scikit-Learn \cite{pedregosa_scikit-learn_2011} implementation with default settings for logistic regression, Naive Bayes and random forest. A Polars variant of Algorithm \ref{alg:CP-python} was applied for the proposed \method{} method. All these methods were executed on a GPU (Amazon EC2 G4 instance). The HGCN was executed using PyTorch-geometric \cite{fey_fast_2019}.

Comparing the execution times in Table \ref{tab:execution-time}, HGCN is the most numerically complex method. Although methods to improve training efficiency, such as batching, are available, they were not considered in this work. Logistic regression and random forest exhibit comparable or shorter execution times in Corona and Movies datasets compared to \method{} and Naive Bayes, largely because the most challenging part—extraction of structural information—is handled by nonnegative matrix factorization, which is not included in the reported times. The proposed \method{} excels particularly in graphs with a low average degree of nodes, similar to Naive Bayes. The measured execution time of \method{} aligns with its expected asymptotic complexity \ref{tab:asymptotic-complexity} and appears to be linear with the number of \method{} layers, as anticipated.

 In summary, HGCN is by far the most numerically intensive method and shows potential in some examples; however, there is still a significant amount of fine-tuning needed to achieve superior performance across datasets. NMF-based methods work exceptionally well on the Movies dataset for the retrieval task, though further tuning is required to properly extract structural information. Compared to Naive Bayes, \method{} is even slightly simpler and parameter-free. In some problems, \method{} outperformed Naive Bayes, and vice versa. Thus, it is definitely a good idea to consider both \method{} and Naive Bayes when establishing baselines for tasks on structural data.

%% file: src/conclusion.tex
\section{Conclusion}

This paper presents a signal propagation algorithm for hypergraphs, termed \methodlong{} (\method{}). We formally describe the \method{} algorithm and demonstrate its simplicity and efficiency of implementation. This formal description allows us to show clear relationships between \method{} and well-known algorithms such as Naive Bayes, label propagation, and hypergraph convolutional networks. These relationships suggest various algorithmic variants, which we left for detailed future exploration.

We discuss the application of \method{} to different types of signals. Our primary focus is propagating binary labels, which is used for classification and retrieval tasks, positioning \method{} as a hypergraph variant of traditional label propagation. Additionally, propagating node features instead of labels as signals leads to feature propagation. This dual functionality showcases the versatility of \method{} in handling various tasks on hypergraphs.

Our evaluation of \method{} on several real-world datasets from multiple domains demonstrates its competitive performance in node classification and retrieval tasks compared to a range of reference methods. Furthermore, we compare the numerical complexity of these methods by examining their execution times, highlighting the simplicity and efficiency of \method{}. This combination of competitive performance and low computational complexity makes \method{} a promising approach for applications involving hypergraph-based data.

%% file: src/credits.tex
\begin{credits}

\subsubsection{\ackname}
This work was supported by the Grant Agency of the Czech Technical University in Prague, grant No. SGS20/183/OHK4/3T/14. Large language models were employed for proofreading, drafting, and clarifying the text; however, the authors always validated and take responsibility for the final result.

\subsubsection{\discintname}
All authors are currently employed by Cisco Systems, Inc. and conducted research presented in this work as part of their job. Pavel Procházka and Lukáš Bajer also own stock in Cisco Systems, Inc. The authors declare that their employment and stock ownership does not influence the objectivity of the presented research.

\end{credits}